\documentclass{article}
\usepackage{subfig}
\usepackage{multirow,multicol}
\usepackage{makecell}
\usepackage{amsfonts}
\usepackage{color}
\usepackage{enumitem}
\usepackage{array}
\usepackage[pagebackref=true,breaklinks=true,letterpaper=true,colorlinks,bookmarks=false]{hyperref}
\usepackage{spconf,amsmath,graphicx}
\newcolumntype{H}{>{\setbox0=\hbox\bgroup}c<{\egroup}@{}}

\def\eqref#1{(\ref{eq:#1})}
\def\eqlabel#1{\label{eq:#1}}
\def\figref#1{\ref{fig:#1}}
\def\figlabel#1{\label{fig:#1}}
\def\pparagraph#1{\par{\bf #1}~~}

\def\eqref#1{(\ref{eq:#1})}
\def\eqlabel#1{\label{eq:#1}}
\def\figref#1{\ref{fig:#1}}
\def\figlabel#1{\label{fig:#1}}
\def\pparagraph#1{\par{\bf #1}~~}

\def\xcomment#1{\textcolor[rgb]{.3,.3,.1}{\text{$/\!\!/$ {\em #1}}}}
\def\comment#1{\kern-1cm\xcomment{#1}}
\def\eqcomment#1{\kern-1cm\xcomment{#1}}

\def\m#1{\ensuremath{\mathtt{#1}}}

\def\v#1{\ensuremath{\mathbf{#1}}}

\def\Rmx#1#2{{\mathbb R}^{{#1}\times{#2}}}

\def\real{\mathbb{R}}
\def\tr{^\top}

\def\norm#1{\left\lVert#1\right\rVert}

\def\l2#1{\norm{#1}_2}

\def\BibTeX{{\rm B\kern-.05em{\sc i\kern-.025em b}\kern-.08em
		T\kern-.1667em\lower.7ex\hbox{E}\kern-.125emX}}

\title{A 3D model-based approach for fitting masks to faces in the wild}
\name{
    \begin{tabular}{@{}c@{}}
    Je Hyeong Hong$^{\star1}$, 
    Hanjo Kim$^{\star2}$, 
    Minsoo Kim$^{1,3}$,
    Gi Pyo Nam$^{1}$, \\ 
    Junghyun Cho$^{1}$, 
    Hyeong-Seok Ko$^{2}$, 
    Ig-Jae Kim$^{1,3}$
    \end{tabular}
    \vspace{-1mm}
    \thanks{$*$ Je Hyeong Hong and Hanjo Kim contributed equally to this work.}
}
\address{
    $^1$Korea Institute of Science and Technology \qquad
    $^2$Seoul National University \qquad \\
    $^3$University of Science and Technology
}
\begin{document}
%
\maketitle
\begin{abstract}

Face recognition now requires a large number of labelled masked face images in the era of this unprecedented COVID-19 pandemic.
Unfortunately, the rapid spread of the virus has left us little time to prepare for such dataset in the wild.
To circumvent this issue, we present a 3D model-based approach called \emph{WearMask3D} for augmenting face images of various poses to the masked face counterparts.
Our method proceeds by first fitting a 3D morphable model on the input image, second overlaying the mask surface onto the face model and warping the respective mask texture, and last projecting the 3D mask back to 2D.
The mask texture is adapted based on the brightness and resolution of the input image.
By working in 3D, our method can produce more natural masked faces of diverse poses from a single mask texture.
To compare precisely between different augmentation approaches, we have constructed a dataset comprising masked and unmasked faces with labels called \emph{MFW-mini}.
Experimental results demonstrate WearMask3D\footnotemark produces more realistic masked faces, and utilizing these images for training leads to state-of-the-art recognition accuracy for masked faces.
\footnotetext{\url{https://github.com/jhh37/wearmask3d}}
\end{abstract}
\section{Introduction}
\label{sec:introduction}

Many countries are imposing a strict mask-wearing policy in public places with aims to prevent transmission of the novel coronavirus through respiratory droplets.
Consequently, face recognition in unconstrained environments has become a rather more difficult task both for humans and machines, each having to identify people through occlusions from their hairstyles, eyes and eyebrows only.
While it is hoped we will one day return to our lifestyle before the COVID-19 pandemic, improving face recognition models to work  with masked faces is a current challenge we are directly facing.

The state-of-the-art face recognition methods~\cite{schroff15,liu17,deng19} are mostly, if not all, data-driven, which have been made possible by the availability of large-scale datasets of labelled face images in the wild, e.g. VGGFace~\cite{omkar15,cao18}, CelebA~\cite{liu15} and CASIA-Webface~\cite{yi14} just to name a few.
It is similarly anticipated a recognition model for masked faces will require a large number of labelled masked face images.
In addition, the identities of masked faces would need to overlap with those of non-masked faces for the models to learn the correct representation of the individuals.
Unfortunately, wearing a mask has only become a common global practice in recent months, and thus it still requires some time for such large-scale masked dataset with labels to be constructed and publicly released.

\begin{figure*}[t]
	\center
	\includegraphics[width=0.95\linewidth]{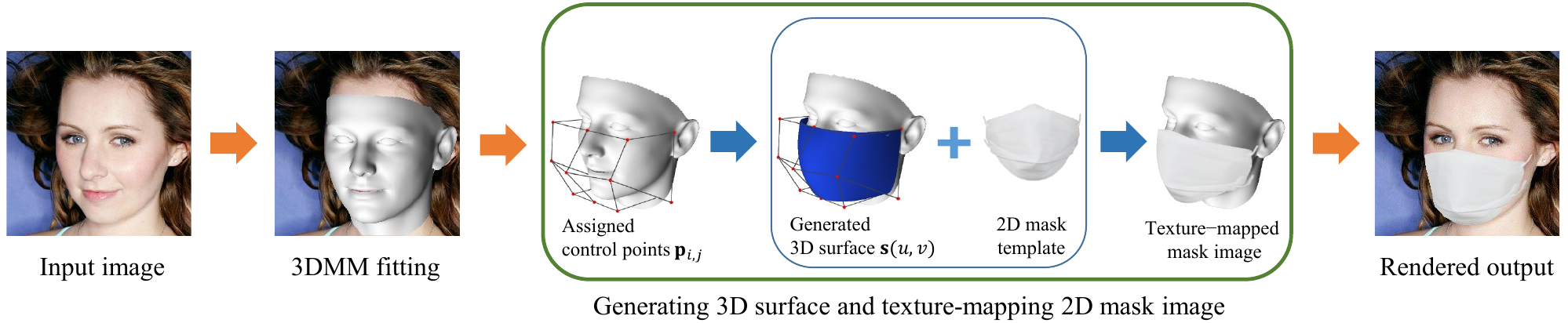}
	\vspace{-4mm}	
	\caption{
		An overview of our \emph{WearMask3D} pipeline.
		We first fit a 3D morphable model (Basel Face Model '09~\cite{paysan09}) using a pretrained off-the-shelf neural 3DMM model regressor called 3DDFA~\cite{zhu19}.
		Then, we compute 15 control points at predefined locations with respect to the 3D model.
		From these control points, we fit a mask surface based on non-uniform rational B-spline (NURBS) and warp the mask texture onto it.
		The 3D warped mask is then projected back to 2D for image rendering.
	}
	\figlabel{pipeline}
	\vspace{-3mm}	
\end{figure*}

Meanwhile, to avoid lack of data from being the bottleneck of research in masked face recognition, a practical short-term solution has been to augment publicly available face datasets to create pseudo-real masked face images.
Ud din et al.~\cite{uddin20} randomly placed frontal mask images to frontal faces using Adobe Photoshop, but this involves a manual process.
Cabani et al.~\cite{cabani21} used 2D facial landmarks to fit 2D masks to faces, but the fitting is natural for near-frontal images only. 
Anwar and Raychowdhury~\cite{anwar20} improved this by utilizing 3 textures (front, left and right) for each type of mask to handle side poses, but this requires additional efforts for each newly added mask and the fitting result is still inaccurate for extreme poses (see Fig.~\figref{qualitative_comparison}).
Automatically fitting masks more naturally to faces of extreme poses serves as our main goal.

In this work, we propose a 3D model-based approach called \emph{WearMask3D} for masking face images in the wild.
Our method utilizes the well-known 3D morphable model (3DMM)~\cite{blanz99,paysan09} to estimate the face surface, from which the mask surface is constructed and the corresponding mask texture warped.
The output is formed by back-projecting the 3D mask to the input image then adjusting the mask brightness and resolution.
The whole pipeline can be processed efficiently as 3DMM coefficients can be regressed using a neural network~\cite{yin17,zhu19} without iterative optimization.

The main contributions of this work are as follows:
\vspace{-3mm}
\begin{itemize}[leftmargin=*]
    \setlength\itemsep{-0.3em}
    \item[+]
    \emph{WearMask3D}, an efficient mask-fitting pipeline based on the 3D morphable model and head pose estimation to produce natural masked faces across a wide range of poses, 
    \item[+]
    \emph{Masked Faces in the Wild (MFW) mini}, a dataset of labeled masked and non-masked faces
    larger than previously reported and allowing various types of verification pairs, and
    \item[+]
    comparisons of WearMask3D against baseline and  state-of-the-art augmentation method (MaskTheFace).
\end{itemize}

\section{Proposed mask augmentation method}
\label{sec:method}

We now illustrate each stage of our pipeline from Fig.~\figref{pipeline}.

\vspace{-2mm}
\subsection{Fitting a 3D morphable model (3DMM)}
\label{sec:3dmm}
The origin of 3DMM dates back to the seminal work of Blanz and Vetter~\cite{blanz99}, in which a 3D face comprising $N$ 3D points, $\m S\in\Rmx{3}{N}$, is expressed as a weighted sum of low-rank basis shapes.
These shapes are grouped into those based on the person's identity and others on the person's expression. 
i.e.
\vspace{-2mm}
\begin{align}
\m S(\boldsymbol{\alpha}_\mathrm{id}, \boldsymbol{\alpha}_\mathrm{exp}) & = \Bar{\m S} + \sum_{j=1}^{N_\mathrm{id}} \alpha_\mathrm{id}^j \m A_\mathrm{id}^j + \sum_{k=1}^{N_\mathrm{exp}} \alpha_\mathrm{exp}^k \m A_\mathrm{exp}^k,
\end{align}
\vspace{-2mm}

where $\Bar{\m S}\in\Rmx{3}{N}$ is the mean shape, $\m A_\mathrm{id}^j\in\Rmx{3}{N}$ is the $j$-th identity basis, $\m A_\mathrm{exp}^k\in\Rmx{3}{N}$ is the $k$-th expression basis, and $\alpha_\mathrm{id}^j$ and $\alpha_\mathrm{exp}^k$ are the scalar weights for the respective bases.

One often uses a pretrained basis model $(\{\m A_\mathrm{id}^j\}, \{\m A_\mathrm{exp}^k\})$, and we also use Gerig et al.'s Basel face model 2009~\cite{gerig18}.

Assuming the face is relatively flat and far away from the camera, each 3DMM point can be projected to 2D using the weak-perspective camera model~\cite{hartley03} as follows:
\vspace{-1mm}
\begin{align}
V(\boldsymbol{\alpha}_\mathrm{id}, \boldsymbol{\alpha}_\mathrm{exp}, \v p) = f \begin{bmatrix} 1 & 0 & 0 \\ 0 & 1 & 0 \end{bmatrix} \m R(\v q) \m S + \v t \v 1\tr,
\eqlabel{model_projection}
\end{align}
\vspace{-1mm}
where $f\in\real$ is the focal length, $\m R\in\,SO(3)$ is the rotation matrix, $\v t\in\real^2$ is translation, $\v q\in\real^4$ are the quaternions representing $\m R$ and $\v p$ denotes the camera variables $[\v f, \v q, \v t]$.

We employ Zhu et al.'s end-to-end pretrained network called 3DDFA~\cite{zhu19} to jointly estimate $\v p$, $\boldsymbol{\alpha}_\mathrm{id}$ and $ \boldsymbol{\alpha}_\mathrm{exp}$ from the input image.
In this case, the number of identity bases ($N_\mathrm{id}$) and expression bases ($N_\mathrm{exp}$) are 40 and 10 respectively, with the number of dense 3DMM points $N$ set to 53,490. 

\vspace{-2mm}
\subsection{Generating 3D mask surface}
\label{sec:3d_mask_surface}
After fitting the 3DMM, we generate a 3D mask surface on which the 2D mask template is mapped. 
For this purpose, We use the 3D non-uniform rational B-spline (NURBS) surface, which is defined for 0$\leq$$u$$\leq$1 and 0$\leq$$v$$\leq$1 as the tensor product of a horizontal and vertical NURBS curves:
\vspace{-1mm}
\begin{align}
\v s(u,v)&= \sum_{i=1}^{K}\sum_{j=1}^{L}\Phi_{i,j}(u,v)\v c_{i,j},
\end{align}
\vspace{-1mm}
where $\Phi_{i,j}(u,v)$ is the rational basis function defined as
\vspace{-1mm}

\begin{align}
\Phi_{i,j}(u,v) &:= \frac{B_{i,n}(u) B_{j,m}(v)w_{i,j} }{\sum_{p=1}^{K}\sum_{q=1}^{L} B_{p,n}(u) B_{q,m}(v)w_{p,q}},
\end{align}
$\v c_{i,j}\in\real^3$ is the $(i,j)$-th control point, $K$ and $L$ are the number of control points for the horizontal and vertical NURBS curves respectively, $B:\real\rightarrow\real$ represents the B-spline basis function, $n$ and $m$ are the polynomial degrees of the NURBS curves, and $w_{i,j}$ is the weight of the control point $(i,j)$.

In our implementation based on NURBS-Python~\cite{bingol2019geomdl}, we assign 15 control points ($K$=3 and $L$=5) for the NURBS surface, second order polynomials for both curves ($m$=2, $n$=2) and equal unit weights across all control points. 
The location of each control point is automatically calculated based on the positions of several 3D facial landmarks such as tip of the nose, cheeks, bottom of the lips in the 3DMM.
Nevertheless, we did go through some trial-and-errors to find the optimal position of each control point for realistic mask-fitting.
The control points are set differently for different types of masks.

\vspace{-2mm}
\subsection{Mapping the mask texture}

We have obtained several mask textures by gathering high-quality redistributable images of  real masked faces in frontal view, cropping the mask region and resizing the image.
The deleted region is left transparent as shown in Fig.~\figref{mask_templates}.

From the frontal (opposite-the-face) viewpoint, we assign $(u,v)= (0,0)$ to the bottom-left vertex of the 3D mask surface and $(u,v)=(1,1)$ to the top right vertex. 
Then, the $uv$-coordinates of approximately 1000 mask surface vertices are obtained through interpolation.

We then arbitrarily select one mask template from Fig.~\figref{mask_templates} (with different probabilities)  and map its texture to the 3D mask surface.
We apply bilinear interpolation to fill in the texture between adjacent surface points.
Since the 3D mask surface smoothly covers the lower half of the face, the mapped mask surface looks realistic in 3D as shown in Fig.~\figref{pipeline}.

\begin{figure}[t]
	\center
		\includegraphics[width=0.95\linewidth]{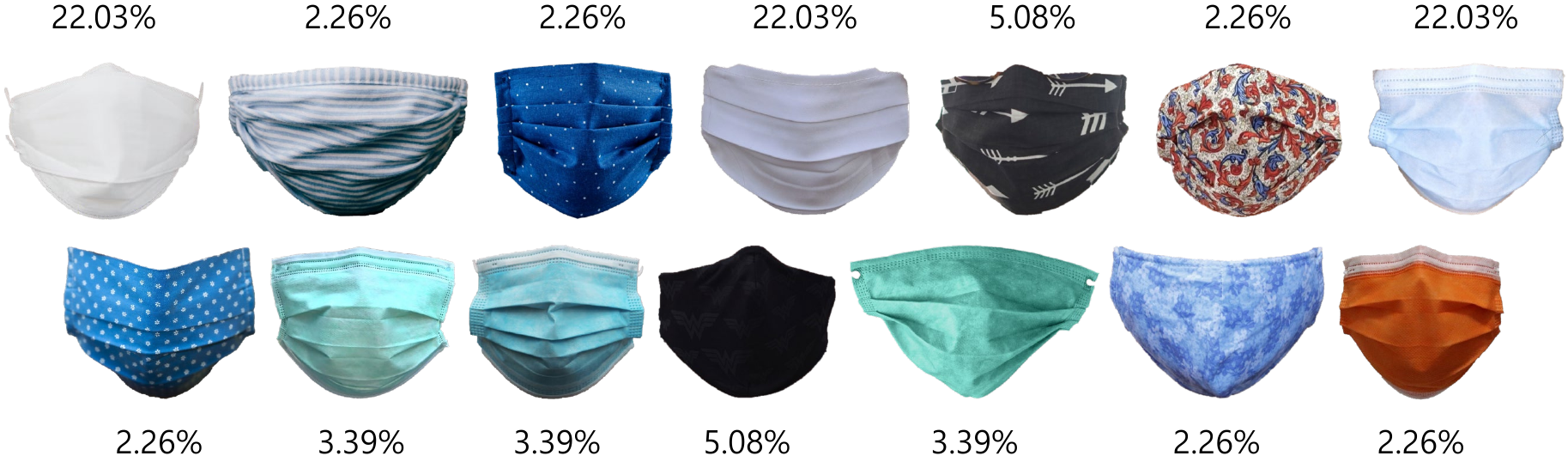}
	\vspace{-3mm}			
	\caption{
		A list of utilized mask templates with probabilities
	}
	\vspace{-6mm}	
	\figlabel{mask_templates}
\end{figure}
\begin{figure}[t]
	\center
	\subfloat[][Lower brightness]{
		\includegraphics[width=0.475\linewidth]{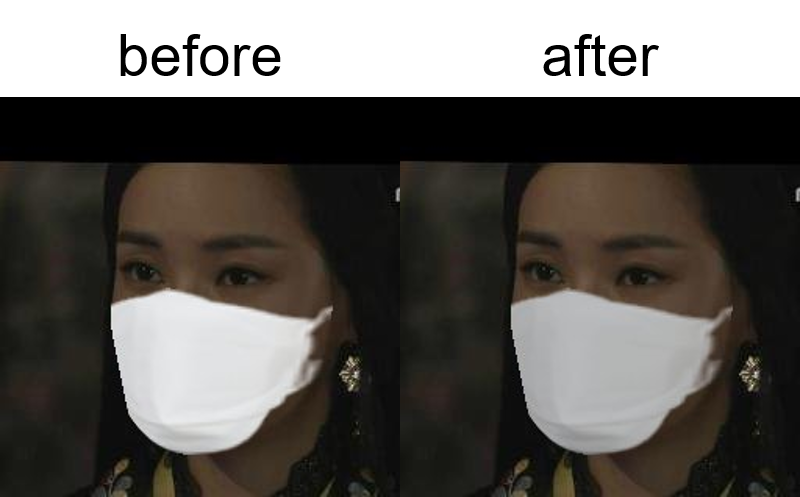}
	}
	\subfloat[][Downsampled mask]{
		\includegraphics[width=0.475\linewidth]{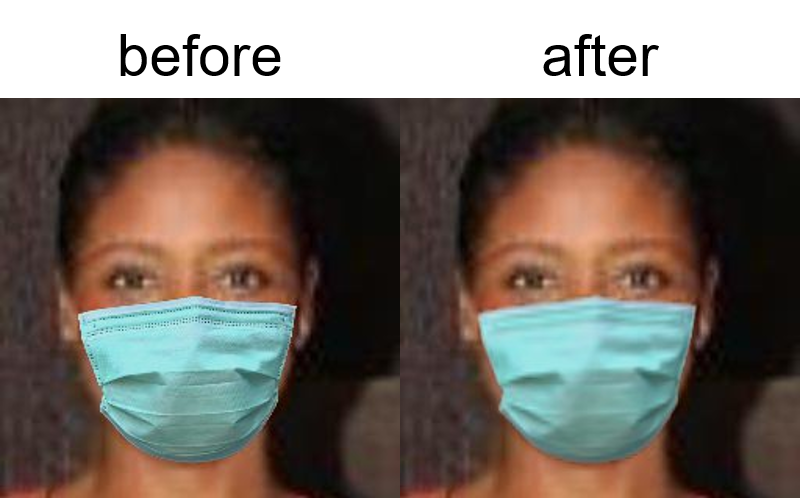}
	}
	\vspace{-2mm}
	\caption{
		Effects of controlling mask brightness and resolution
	}
	\figlabel{brightness_resolution}
	\vspace{-4mm}
\end{figure}

\vspace{-2mm}
\subsection{Rendering and post-processing}
Finally, we render the texture-mapped 3D mask surface on the 2D input image using the camera pose obtained using 3DDFA in Sec.~\ref{sec:3dmm}. 
So long as the 3DMM is accurately fitted on the input image, the rendered mask also fits the face naturally.
We then refine the mask rendering for quality improvement.

\pparagraph{Brightness control}
Just rendering the mask with equal brightness across all images sometimes produces visually unsatisfactory results largely due to significant variations of illumination across images. 
To mitigate this issue, we estimate the brightness of the input image by averaging the intensities (obtained by converting RGB to greyscale and dividing by 255) of the entire pixels. 
If the measured brightness is lower than our pre-determined threshold of 0.4, we reduce brightness of the mask $\rho$ as $0.7 + 0.75*\min(0.4,\rho)$, where the numbers are deduced from empirical trials.

\begin{figure}[t]
	\center
	\includegraphics[width=0.95\linewidth]{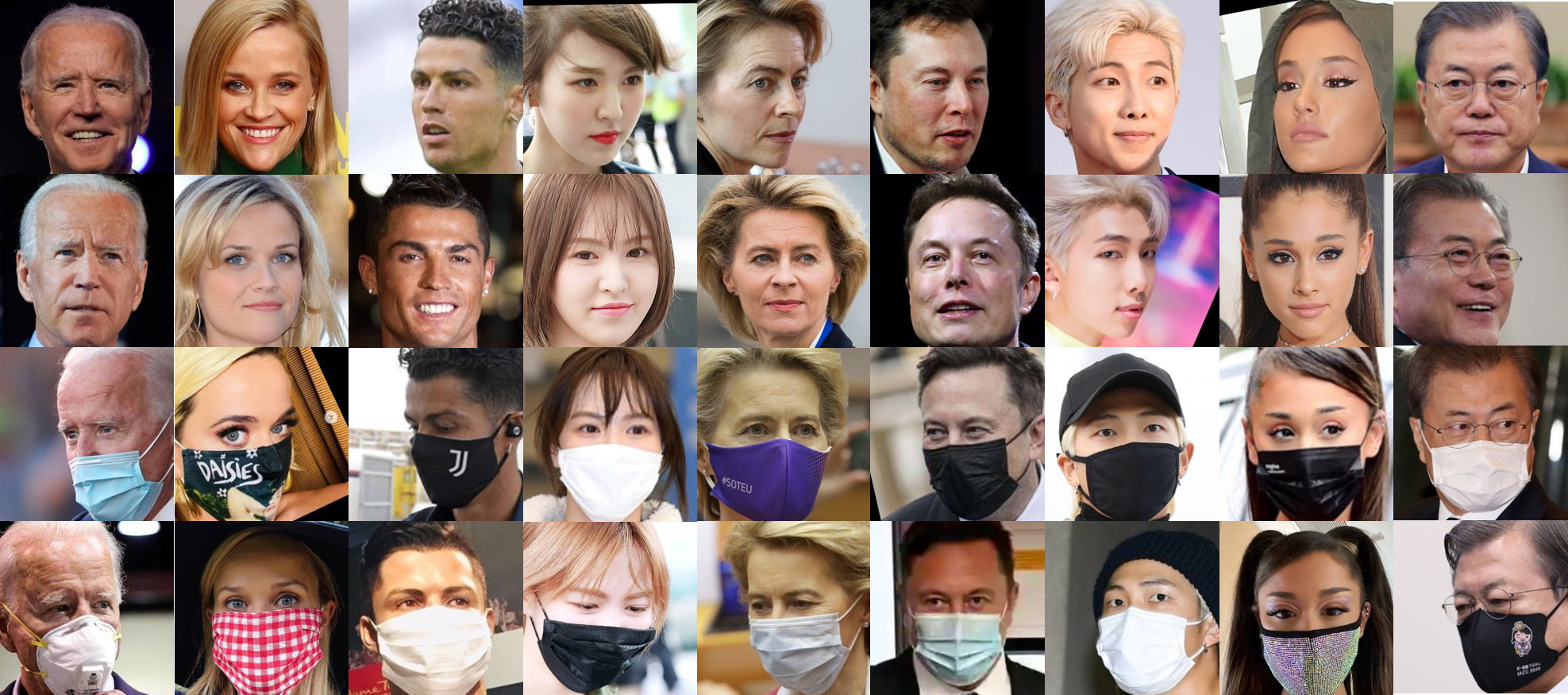}
	\vspace{-4mm}	
	\caption{
		Images from our class-balanced \emph{MFW-mini} dataset.
	}
	\vspace{-4mm}
	\figlabel{mfw_mini}
\end{figure}

\pparagraph{Resolution adjustment}
We also adjust the mask resolution depending on the sharpness of the input image. Variance of the Laplacian has been known as an effective concept to estimate the image sharpness \cite{903548}. We apply a conventional $3\times3$ Laplacian kernel to each pixel of the input image and calculate its variance $\sigma^2$.
If $\sigma^2$\,$<$400, we regard the input as a low-quality image and reduce its resolution by the factor of $\min(10, 400 / \sigma^2)$, where the values are again empirically set. 

\begin{figure*}[t]
	\center
	\includegraphics[width=0.97\linewidth]{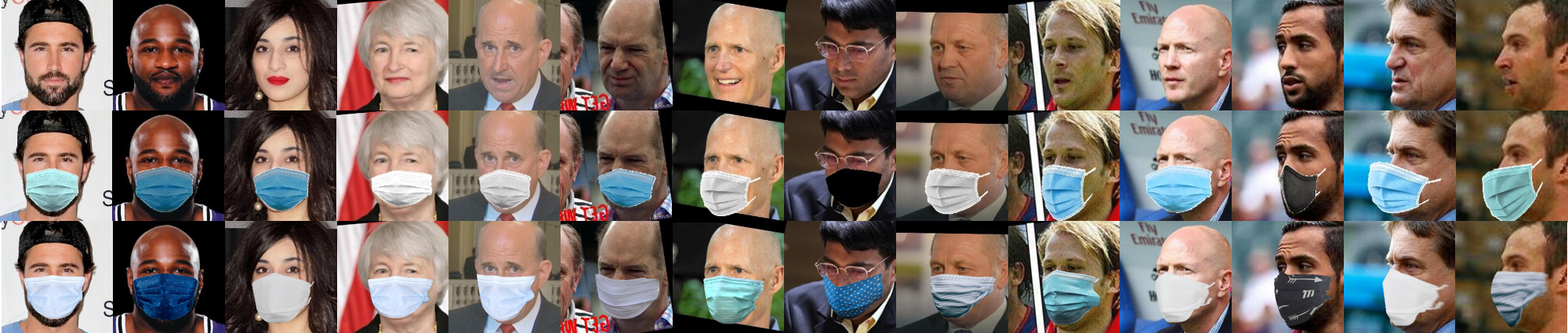}
	\vspace{-4mm}	
	\caption{
		A visual comparison of the masked faces. The top row contains original images from VGGFace2~\cite{cao18}, while the middle and bottom rows show outputs from MaskTheFace~\cite{anwar20} and ours respectively.
		The horizontal pose varies from 0$^\circ$ to 90$^\circ$.
	}
	\figlabel{qualitative_comparison}
	\vspace{-4mm}	
\end{figure*}

\begin{table}[t]
	\centering
	\caption{
		Fr\'{e}chet inception distances (FIDs) achieved by different mask augmentation methods on MFR2~\cite{anwar20} and MFW-mini (lower is better).
		The tested feature dimensions are set to be below the number of images available in each dataset.
	}
	\iftrue
	\begin{tabular}{l | r | r | r | r}
		\hline\hline
		\makecell[l]{Augmentation} & 
		\multicolumn{1}{c|}{\makecell[c]{MFR2}}	&
		\multicolumn{3}{c}{\makecell[c]{MFW-mini (Sec.~\ref{sec:mfw_mini})}}
		\\
		\makecell[l]{method}& 
		\makecell[c]{64-d} & 
		\makecell[c]{64-d} &
		\makecell[c]{192-d} & 
		\makecell[c]{768-d} 		
		\\  \hline
		\makecell[l]{No aug. (non-masked)} & 
		\makecell[r]{0.665} &
		\makecell[r]{0.556} & 		
		\makecell[r]{3.114} &
		\makecell[r]{0.954}
		\\		
		\makecell[l]{MaskTheFace~\cite{anwar20}} & 
		\makecell[r]{0.468} &		
		\makecell[r]{0.410} & 		
		\makecell[r]{2.011} &
		\makecell[r]{0.304} 		
		\\	\hline
		\makecell[l]{\textbf{WearMask3D
		 }} & 
		\makecell[r]{\textbf{0.398}} &		 
		\makecell[r]{\textbf{0.153}} & 		
		\makecell[r]{0.856} &
		\makecell[r]{\textbf{0.189}}	\\	
		\makecell[l]{~~~w/o brightness adj.} & 
		\makecell[r]{0.404} & 		
		\makecell[r]{0.154} &
		\makecell[r]{\textbf{0.854}} & 		
		\makecell[r]{\textbf{0.189}}	\\
		\makecell[l]{~~~w/o resolution adj.} & 
		\makecell[r]{0.399} & 		
		\makecell[r]{0.203} &
		\makecell[r]{1.171} & 		
		\makecell[r]{0.200}	\\		
		\hline\hline	
	\end{tabular}
	\fi
	\label{tbl:fid_scores}
	\vspace{-4mm}	
\end{table}	

\begin{table}[t]
	\centering
	\vspace{-2mm}
	\caption{
		Verification accuracies~(\%@EER) achieved by different mask augmentation schemes.
		N+N denotes pairs of non-masked images, M+N pairs of masked and non-masked images, M+M pairs of masked images, and Mix. mixed pairs.
		``Bottom-out'' overwrites 0s to the bottom 45\% of the image.
	}	
	\begin{tabular}{l | r | r | r | r | r}
		\hline\hline
		\makecell[l]{Augmentation} & 
		\makecell[c]{LFW} &
		\makecell[c]{MFR2} &		
		\multicolumn{3}{c}{MFW-mini (Sec.~\ref{sec:mfw_mini})} \\
		\makecell[l]{Method} & 
		\makecell[c]{-} &
		\makecell[c]{Mix.} &
		\makecell[c]{N+N}& \makecell[c]{M+N} & \makecell[c]{M+M} \\ \hline
		
        \makecell[l]{No augment.} & 
		\makecell[r]{\textbf{97.3}} &  
		\makecell[r]{90.1} & 		
		\makecell[r]{\textbf{96.7}} & 
		\makecell[r]{87.3} & 
		\makecell[r]{89.7} \\
        
		\makecell[l]{Bottom-out} & 
		\makecell[r]{96.3} &  
		\makecell[r]{93.9} & 		
		\makecell[r]{96.1} & 
		\makecell[r]{89.5} & 
		\makecell[r]{91.1} \\		
        
		\makecell[l]{MaskTheFace} & 
		\makecell[r]{96.6} &
		\makecell[r]{94.6} & 		
		\makecell[r]{95.9} & 
		\makecell[r]{90.3} & 
		\makecell[r]{92.1} \\  
        
		\makecell[l]{\textbf{WearMask3D}} & 
		\makecell[r]{96.6} &
		\makecell[r]{\textbf{95.8}} & 		
		\makecell[r]{95.9} & 
		\makecell[r]{\textbf{90.9}} & 
		\makecell[r]{\textbf{92.6}} \\			
		\hline\hline	
	\end{tabular}
	\label{tbl:verification_accuracies}
	\vspace{-5mm}	
\end{table}

\section{Masked faces in the wild with labels}
\label{sec:mfw_mini}
So far, the largest public dataset of real labelled masked faces has been MFR2~\cite{anwar20}, comprising 171 masked and 98 non-masked faces from 53 identities with 848 test pairs for verification.
We found this to be small for precise comparisons, and decided to construct a larger dataset for testing.

We gathered 3,000 images of 300 individuals from the internet, forming a class-balanced dataset by collecting 5 masked and 5 non-masked images per identity.
This is to incorporate various matching situations, e.g. between masked and non-masked faces as well as masked vs masked.
We cropped and aligned images using the bounding box and 5 facial landmarks extracted from RetinaFace~\cite{deng20}.
About 3\% of images were incorrectly cropped or aligned, and these were re-cropped by manually extracting the 5 facial landmarks.
The dataset is named \emph{Masked Faces in the Wild (MFW) mini}, and some of the constituent images are shown in Fig.~\figref{mfw_mini}.

While the number of images is still relatively small compared to that of LFW~\cite{huang07}, this allows 13,500 genuine pairs to be formed (compared to $\approx$400 for MFR2) including 3,000 genuine pairs of masked images, allowing the associated verification accuracies to be more reliable.
We plan to release either the actual dataset or the URLs of the utilized  images.

\section{Experimental results}
\label{sec:results}
Our approach is compared against baseline (no augmentation) and state-of-the-art mask augmentation method (MaskTheFace~\cite{anwar20}) on two aspects, namely the quality of masked images by calculating the Fr\'{e}chet inception distance (FID)~\cite{heusel17} and the usefulness of mask augmentation for training by measuring the accuracy improvement in masked face recognition.

\vspace{-2mm}
\subsection{Image quality analysis}
Fig.~\figref{qualitative_comparison} shows representative mask augmentation results across full range of horizontal poses (0--90$^\circ$). 
This shows visually our method yields more natural mask fitting for profile faces.

We also compared the FIDs of different approaches on MFR2 and MFW-mini.
FID is the Fr\'{e}chet distance between the distribution of the Inception v3  features from real samples and those from fake ones (lower is better).
While FID is frequently used to compare the quality of fake images produced by a deep generator, we would like to stress that FID itself does not enforce any constraint on the nature of fake images, i.e. a fake image can be produced by a 3D rendering process.

On each dataset, we fitted a mask to each non-masked face using MaskedTheFace and our method.
Then, the FID between the real masked faces and the mask-fitted images was calculated for each method.
For the baseline comparison, we also provide the FID between the real masked images and non-masked images.

While the FID is usually computed on 2,048-d Inception v3 features, we had to use lower dimensional features (64, 192 and 768) as the number of non-masked images is smaller than 2,048 for MFR2 (98) and MFW-mini (1,500).
Table~\ref{tbl:fid_scores} shows WearMask3D yields lower FIDs across all tested Inception v3 features, indicating more realistic mask augmentation.

\vspace{-2mm}
\subsection{Masked-face recognition accuracy}

We compared the verification accuracies by training a network with mask augmented images from different methods.

As in~\cite{anwar20}, we created a class-balanced random subset of VGGFace2~\cite{cao18} for training, with 42 images for each of the 8,631 identities.
For each augmentation method, we fitted masks to 362,502 faces and used them with the original non-masked subset for training over 25 epochs.
The ``no augmentation'' setting was trained with the non-masked subset only but for 50 epochs to compensate for halved number of training images.
Across all settings, we used the ResNet-50 backbone, and applied basic softmax loss for training.

We set the initial learning rate to $10^{-1}$, and slowed the learning rate by factor of 10 at specified epochs (16, 28, 40 for ``no augmentation'' and 8, 14, 20 for others).
We set momentum to 0.9 and weight decay to 5e-4 as in~\cite{deng19}. 
We used batch size of 256 for ``no augmentation'' and 512 (256 non-masked + 256 masked) for others, distributed over 8 Nvidia RTX 2080~Ti GPUs.

For evaluation, we utilized authors-provided 848 test pairs for MFR2, and created 6,000 unique test pairs (50\% genuine) for each MFW-mini setting.
For detailed analysis, we varied types of tested face pairs, namely non-masked (N+N), masked and non-masked (M+N), masked (M+M).
In each pair, we extracted 2,048-d feature vectors and calculated the angle between them.
For each method on each setting, we deduced the acceptance threshold $\phi$ at which the equal error rate holds~(FAR=FRR), and reported the sum of true positives and true negatives over the total number of test pairs.

Table~\ref{tbl:verification_accuracies} shows WearMask3D achieves state-of-the-art  recognition accuracy when test pairs include real masked faces.
On the other hand, the network with no augmentation performs best for non-masked face pairs, implying better network training is required to cover all verification cases.

\vspace{-2mm}
\section{Conclusion}
\label{sec:conclusion}
\vspace{-1mm}

In this work, we have proposed \emph{WearMask3D}, an automatic 3D model-based approach for fitting masks to face images in the wild.
We have shown that incorporating the 3D morphable model allows more naturally mask fitted faces even with extreme poses.
We have also demonstrated the recognition accuracy of masked faces can be improved by training with masked faces generated from our approach.
As a by-product, we have created \emph{MFW-mini}, a dataset of 300 identities each with 5 masked and 5 non-masked faces in the wild that can be used for testing.
We have released the relevant code aims to facilitate future research.

\vspace{2mm}
\noindent
\textbf{Acknowledgements }
This work was supported by the R\&D program for Advanced
Integrated-intelligence for Identification (AIID)
through the National Research Foundation of Korea (NRF) funded by the Ministry of Science and ICT
(2018M3E3A1057288), by the Korea Institute of Science and Technology (KIST) project ``Multimodal Visual Intelligence for Cognitive Enhancement of AI Robots", by the Brain of Korea 21 project 2021,
and by the Automation and Systems Research Institute (ASRI) at
Seoul National University.

\bibliographystyle{IEEEbib}
\bibliography{bibliography/mask}

\end{document}